# Open-set Intersection Intention Prediction for Autonomous Driving

Fei Li, Xiangxu Li, Jun Luo, Shiwei Fan and Hongbo Zhang
{lifei120, lixiangxu, jun.luo1, fanshiwei, zhanghongbo888}@huawei.com
Noah's Ark Lab, Huawei Technologies Co., Ltd.

*Abstract*— Intention prediction is a crucial task for Autonomous Driving (AD). Due to the variety of size and layout of intersections, it is challenging to predict intention of human driver at different intersections, especially unseen and irregular intersections. In this paper, we formulate the prediction of intention at intersections as an open-set prediction problem that requires context specific matching of the target vehicle state and the diverse intersection configurations that are in principle unbounded. We capture map-centric features that correspond to intersection structures under a spatial-temporal graph representation, and use two MAAMs (mutually auxiliary attention module) that cover respectively lane-level and exit-level intentions to predict a target that best matches intersection elements in map-centric feature space. Under our model, attention scores estimate the probability distribution of the open-set intentions that are contextually defined by the structure of the current intersection. The proposed model is trained and evaluated on simulated dataset. Furthermore, the model, trained on simulated dataset and without any fine tuning, is directly validated on in-house real-world dataset collected at 98 real-world intersections and exhibits satisfactory performance, demonstrating the practical viability of our approach.

## I. INTRODUCTION

Intention prediction is one of the most challenging task in AD systems. The required method needs to handle intention prediction at diverse intersections, including many uncommon and irregular intersections in the real world. There are various approaches [1-3] that utilize traditional machine learning algorithms, such as Hidden Markov Models (HMM), Support Vector Machines (SVM), and Dynamic Bayesian Network (DBN). Recent works tend to use Deep Neural Networks (DNN) to predict both high level behavior and future trajectories [4-8]. These works usually formulate intention prediction as a *closed-set classification* problem with fixed number of ego-centrically defined intentions, such as turning right, turning left and going straight [1-7]. There is no doubt that the structure of many real-world urban intersections follow a fixed set of two-way or three-way choices at 90 degrees with each other. The closed-set formulation of the problem, however, cannot adequately account for real behavior of human drivers at less common but still abundant intersections in the real world. For example as shown in Figure 1, a turn-left prediction may be simultaneously compatible with more than one exits (Exit2 and 3 in Figure 1), rendering the closed-set classification formulation inappropriate.

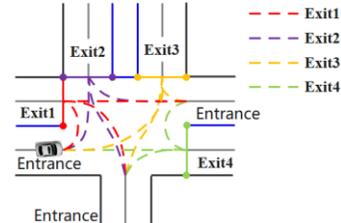

Figure 1. Open-set intersection intention. Open-set intention includes exit or goal level and lane level intention. Each color represents one exit. Each exit consists of a virtual lane set and an exit goal. Each dotted line represents one lane intention. Each colored solid line stands for one goal. Black and blue lines mark road boundary and direction separation.

In this paper, we introduce *open-set intersection intention prediction.* We take the behaviors of human drivers at any intersection to conform to the actual options at that intersection. In other words, we assume that the intention structure, while ego-centrically expressible to some extent, is ultimately only expressible without ambiguity in a map-centric way. Moreover, as illustrated in Figure 1, we recognize that the driver's intention could match the intersection structure at *two levels*: the exit or goal level and the lane level, with each exit goal subsuming under it a set of lanes. As illustrated in Figure 1, one exit goal could be represented by the straight line between the road boundaries at the corresponding exit and the related virtual lane set is the set of virtual lanes that connect to that exit from all the entrances and describe the connectivity available at the intersection. With reference to this general *bi-level* abstraction of intersection structures, we formulate the intersection intention prediction problem as the problem of predicting the intended goal exit and virtual lane. This formulation is *map-centric* in virtue of its reference to the bi-level goal and lane structure based on the map. Moreover, insofar as it introduces no constraints on how many exits and lanes there could be at an intersection, it also treats the problem as an *open-set* bi-level prediction problem.

It should already be clear that this open-set formulation has greater generality than a closed-set formulation, because it is suitable for any intersections with any number of exits or lanes at any angles with each other. In this paper, we show that this open-set bi-level intention prediction problem could be satisfactorily solved by combining a spatiotemporal graph representation with a mutually auxiliary attention mechanism based on Graph Attention Network (GAT) [9-10].

Our theoretical contributions are as follows:

- We use *open-set intentions* to describe behaviors of human driver at any kind of intersections without ambiguity. This formulation eliminates the ambiguity that defeats the closed-set formulation in some scenarios and is thus more versatile and practical in real-world AD systems.
- We further formulate the open-set intention prediction problem as a *probabilistic matching problem* instead of a classification problem. This allows map information to be used directly as structured priors through map-centric representations, instead of having to encode it in high dimensional rasterized images in ego-centric coordinates.
- We use *spatiotemporal graph* and *graph attention mechanisms* to encode and process features in a spatially and permutationally invariant way. This allows our model to generalize to unseen intersections, without any architecture change or model retraining.
- We introduce a novel *mutually auxiliary attention mechanism* that matches estimated target states with map elements at both the exit-goal and lane levels. This simultaneously solves goal intention prediction and lane intention prediction and achieves higher accuracy through both using the level-specific information and exploiting their mutually constraints.

In terms of systems and experiments, we accomplished the following:

- We generated a dataset of simulated trajectories based on 145 intersections in the Shanghai Pudong area with virtual lanes connecting entrances and exits. The 145 intersections include various layouts, such as T-junction, standard four way intersection and irregular intersections with two, three, four even more exits. This simulated dataset is randomly split into training, validation and test set, with 77, 34 and 34 intersections respectively.
- We constructed an in-house real-world dataset of 6346 trajectories collected at 98 intersections from Shanghai Pudong, of which 47 intersections are not in the simulated training set.
- We implemented and trained the proposed model on the simulated training set and evaluated the trained model on both the simulated test set and the in-house dataset. Evaluation results show that our model generalizes to unseen intersections with high performance and achieves 95.9% average recall on the in-house dataset, 95.1% at the unseen intersections.
- The average processing time for each target is about 0.3ms, which shows that our model is suitable for deployment on vehicle platforms for real-world use.

This paper is organized as follows. Some related works is introduced in Section II. The proposed model is presented in detail in Section III. Experiment results and conclusions are given in Section IV and Section V respectively.

## II. RELATED WORK

Intention prediction is a crucial task for navigation and planning [11], and it is useful in ADAS features such as Intersection Movement Assist (IMA) and Left Turn Assist (LTA) and in autonomous driving applications in urban area.

Most existing works predefine a closed-set of intentions offline. For highways or regular road in urban area, the predefined intentions usually consists of left lane change, right lane change and lane keeping [12-13]. For intersections, there are turning left, turning right, and going straight [1-5]. In other words, most works formulate intention prediction as a closed-set classification problem. In [2-3], techniques such as HMM, SVM and DBN are used as classifier with some base features. Recent works mainly use DNN models for intention prediction. The prediction module in Apollo [4] uses rasterized images to represent map information and past states of the target vehicle, and combines CNNs and MLPs to classify the target's intention into three semantic intentions. IntentNet [5] predicts intention from past and current raw sensor data and high dimensional rasterized map data using Convolutional Neural Networks (CNNs), with 8 classes: keep lane, turn left, turn right, left change lane, right change lane, stopping/stopped, parked and other. Deo et al [6] uses social pooling layers to encode data of surrounding vehicles on a highway, and predict the target's maneuver. MultiPath [7] leverages a fixed set of future trajectory anchors corresponding to different intentions, and utilizes CNNs to predict a discrete distribution over the predefined anchors.

Different from above works with predefined intentions, TPNet [8] predicts a rough end point with reference lane, generates a candidate set of future trajectories as hypothesis proposals, and then makes the final predictions by refining the proposals. The rough end point or its related reference lane can be viewed as intention.

Furthermore, these existing works almost invariably require current and past states of target vehicle on hand and rasterize map data as high dimensional tensor [4-5, 7-8]. This leads to a computationally heavy pipeline that gives rise to longer response time.

A commonly used dataset for intention prediction is the NGSIM dataset [15-16]. This dataset mainly focus on highway prediction, and contains only nine intersections. The dataset used in [16-18] was collected from only one roundabout. In [2], its dataset includes only 6 intersection. In contrast, our simulated dataset is generated from 145 different intersections in Shanghai Pudong area, our in-house real-world dataset is collected at 98 intersections of these 145 intersections.

## III. METHODOLOGY

### A. Overview

As mentioned before, we formulate intersection intention prediction as a problem of matching the predicted target with map elements at the exit and lane levels. Given encoded map-centric features at current time and previous hidden states of

RNNs, we use two MAAMs to calculate matching scores for each level intention, and choose the goal with the highest score as the intended goal and the lane with the highest score as the intended lane. Under this architecture, our proposed model can predict intention at arbitrarily complex intersections, so long as their structure can be adequately captured by the open-set intention representation as illustrated in Figure 1.

### B. Spatiotemporal Representation

To represent intersection structures from the map and their relations with target vehicle, we use spatiotemporal graphs illustrated in Figure 2, where temporal edges are omitted for brevity. Each exit goal or virtual lane is represented by a node. Thus, for each exit, we include in the graph a single goal node and multiple virtual lane nodes under that exit. The target vehicle is also a node connected to other nodes. We associate map-centric features with the edges in the graph. RNNs and attention mechanism are used to propagate information in this spatiotemporal graph, which models the interactions between the target vehicle and the intersection structure.

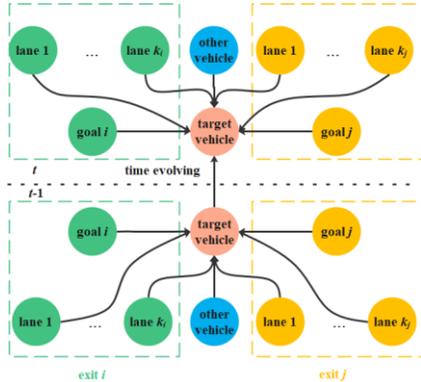

Figure 2. Spatial-temporal graph, each color represents one exit consisting of related lanes and goal. $k_j$ denotes lane number of the $j$-th exit.

### C. Model Architecture

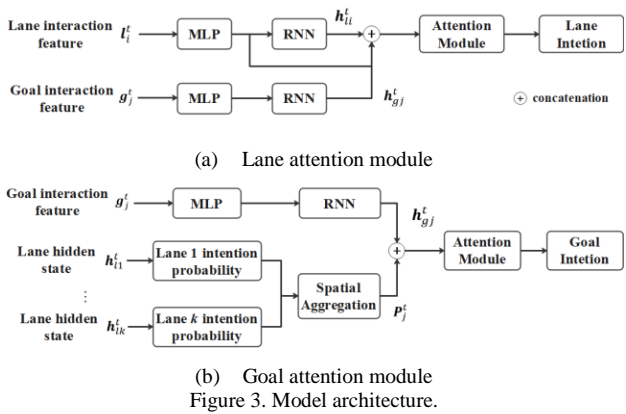

(a) Lane attention module

(b) Goal attention module

Figure 3. Model architecture.

Architecture of the proposed model is illustrated in Figure 3. The proposed model includes a feature encoding module, a temporal propagation module, and two MAAMs [10]. Map-centric features are used to model the interactions between target vehicle and map elements. RNNs and MLPs are combined to propagate information between adjacent frames, then two MAAMs are used to aggregate information from connecting edges. Specifically, the (a) Lane attention module, with the aid of goal information $g_j^t$, computes the attention score of each lane that is normalized to a distribution of lane intentions; and the (b) Goal attention module computes the lane info $P_j^t$ through a weighted sum of the $j$-th exit's related lanes' hidden states $h_{li}^t$, and then concatenates $P_j^t$ with goal hidden state $h_{gj}^t$ to compute attention score of the $j$-th goal. Different lane interaction features share the same lane-specific MLP and RNN. Different goal interaction features also share the same goal-specific MLP and RNN. $l_i^t$ and $g_j^t$ respectively denote the $i$-th lane and the $j$-th goal interaction feature at the $t$-th time.

### D. Feature Encoding

Map-centric features can describe whether the target vehicle is approaching a goal or a lane, these are used as the feature values associated with the spatial edges and can be taken to represent interaction between target vehicle and map elements. Map-centric features include lane interaction feature $l_i^t$ and goal interaction feature $g_j^t$, where $i$ and $j$ are the index of virtual lane and goal and $t$ denotes the $t$-th timestamp. Coordinate systems used here are illustrated in Figure. 4. $l_i^t$ consists of ($s$, $d$) coordinates in a 2-d curvilinear coordinate system with the axes tangential and perpendicular to the lane centerline, lane-relative heading, and differences of above features between adjacent frames. Similarly, a goal coordinate system is defined with origin as the middle point of its corresponding exit, axes tangential and perpendicular to its traffic direction. $g_j^t$ consists of the vehicle's coordinates in the goal coordinate system, heading, distance to origin, and also the differences of above features.

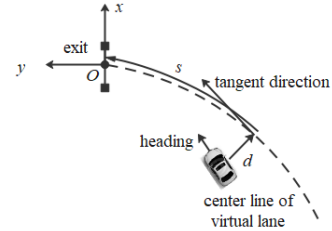

Figure. 4 Coordinate systems used in feature extraction. ($s$, $d$) is the 2-d curvilinear coordinate system, $s$ and $d$ are the tangential and normal distances. ($x$, $y$) is the goal coordinate system. $O$ is the origin of both coordinate systems.

We should note that the combination of our spatiotemporal graph representation and map-centric feature encoding allows our model to be applicable regardless of how the elements of the intersection are located or combined. Because all relevant transformations are captured in an element-specific way, insofar as the DNN generalizes, our model can function in a way that is spatially and permutationally invariant. We believe this is the reason for its great generalization power.

### E. Temporal Propagation Module

We use RNNs to propagate information of temporal edges. The map centric features are embedded by a multi-layer perceptron (MLP) first, and then encoded through a gated recurrent unit (GRU) cell to model the temporal evolution, which are simpler compared to LSTMs and, achieve similar performance here.

$$e_{li}^t = MLP_l(l_i^t) \quad (1)$$
$$e_{gj}^t = MLP_g(g_j^t) \quad (2)$$
$$h_{li}^t = GRU_l(h_{li}^{t-1}, e_{li}^t) \quad (3)$$
$$h_{gj}^t = GRU_g(h_{gj}^{t-1}, e_{gj}^t) \quad (4)$$

where $e_{li}^t$ and $e_{gj}^t$ are the embedding vectors of lane interaction feature and goal interaction feature, and $h_{li}^t$ and $h_{gj}^t$ are the hidden states.

### F. Mutually Auxiliary Attention Module

Lane intention and goal intention are always correlated with each other. For example, if target vehicle take the *j*-th exit to leave intersection, it definitely takes one of the lanes related to the *j*-th exit's to go through the intersection. Accordingly, the proposed model is trained in an end-to-end fashion with the two attention modules facilitating each other. We call them mutually auxiliary attention modules (MAAMs), whose respective lane attention and goal attention scores are probability distributions of corresponding intentions.

**Lane Attention:** For each virtual lane, current lane hidden state $h_{li}^t$, lane embedding vector $e_{li}^t$ and its related goal hidden state $h_{gj}^t$ are concatenated together, then attention score is calculated by an MLP. Related goal hidden state $h_{gj}^t$ is used as auxiliary information to improve the accuracy of lane intention prediction. Formally,

$$\alpha_{ji} = \frac{exp\left(MLP([h_{gj}^t, h_{li}^t, e_{li}^t])\right)}{\sum_{m,k \in set_m} exp\left(MLP([h_{gm}^t, h_{lk}^t, e_{lk}^t])\right)} \quad (5)$$

where $[\cdot]$ is the concatenation operator, $h_{gm}^t$ is the goal hidden state of the *m*-th goal, $set_m$ is the lane set of the *m*-th exit's related lanes, and the softmax operation is conducted across all virtual lanes in intersection. $\alpha_{ji}$ is the intention probability of the *i*-th virtual lane belonging to the *j*-th exit.

**Goal Attention:** Spatial aggregation is defined in equation (6) to obtain the overall representation $P_j^t$ of the *j*-th exit's related virtual lanes. With consideration of $P_j^t$, goal attention $\beta_j$ is computed by a similar attention layer. Related lane representation $P_j^t$ is the detailed interaction information and can improve the robustness to the noise in the map information, such as the absence of centerlines. $\beta_j$ is the intention probability of the *j*-th goal, and it can describe how possibly the target vehicle approach the *j*-th goal. So far, we can see goal intention is predicted separately for each goal.

$$P_j^t = \sum_{i \in set_j} \alpha_{ji} h_{li}^t \quad (6)$$
$$\beta_j = softmax\left(MLP([h_{gj}^t, P_j^t])\right) \quad (7)$$

### G. Loss Functions

During training, we minimize a multi-task objective as the following loss functions.

$$L = w_{lane} L_{lane} + w_{goal} L_{goal} \quad (8)$$

where $L_{lane}$, $L_{goal}$ are the losses for lane intention and goal intention prediction respectively, with $w_{lane}, w_{goal}$ being the corresponding weighting hyper parameters.

### H. Implementation Details

MLPs and GRU cells in temporal propagation have 64 and 128 hidden units respectively. $L_{goal}$ is set to binomial cross entropy loss with weight of positive samples being 4, $L_{lane}$ is set to cross entropy loss. Loss weights $w_{lane}, w_{goal}$ are set to 1, 1. The model is trained using the Adam optimizer with an initial learning rate of 0.001 for 50 epochs, with batch size of 512 and learning rate is exponentially decayed every 10 epochs with a coefficient of 0.9. The hyper parameters were chosen empirically in this work.

## IV. EXPERIMENTS

### A. Dataset

We evaluate our model on a simulated dataset and an in-house real-world dataset. These datasets share the same HD maps, which includes 145 intersections. In the simulated dataset, initial velocity and acceleration of the target vehicle are randomly sampled in a reasonable range, trajectories are generated according to initial velocity, acceleration and reference lane, which is similar to the proposal generation in [8]. Thus, the simulated dataset contains the labels of lane intention and goal intention. According to the angle difference between their two ends, trajectories are classified into two categories: straight trajectory and curved trajectory. Simulated dataset is split into training, validation and test sets. Details are shown in Table I.

TABLE I. DETAILS OF SIMULATED DATASET

| Simulated dataset | *Straight trajectory* | *Curved trajectory* | *Number of intersections* | *Number of sequences* |
|---|---|---|---|---|
| Training set | 2352 | 3470 | 77 | 89000 |
| Validation set | 695 | 1293 | 34 | 29000 |
| Test set | 708 | 1380 | 34 | 30000 |

The in-house dataset is collected at 98 intersections, which is a subset of the total 145 intersections, and 47 of these were not included in generating the simulated *training* set. It contains trajectories of real vehicles estimated with an in-house perception system and manually labeled goal intention. The total number of trajectories is 6346, and the average length of trajectories is about 5 seconds. The class imbalance is significant with 80% straight and 20% curved trajectory. In the in-house dataset, we do not have the label of lane intention. The HD map includes centerline of each lane. Trajectories are sampled at 25Hz in both datasets.

### B. Baseline Methods

**KNN:** Two k-Nearest Neighbor classifiers are used to rank lanes and goals, *k* is set to 9 in both classifiers. Goal and lane interaction features at current time are used for goal intention and lane intention prediction respectively.

**MLP:** MLPs are used to predict lane intention and goal intention separately, batch normalization and the Rectified

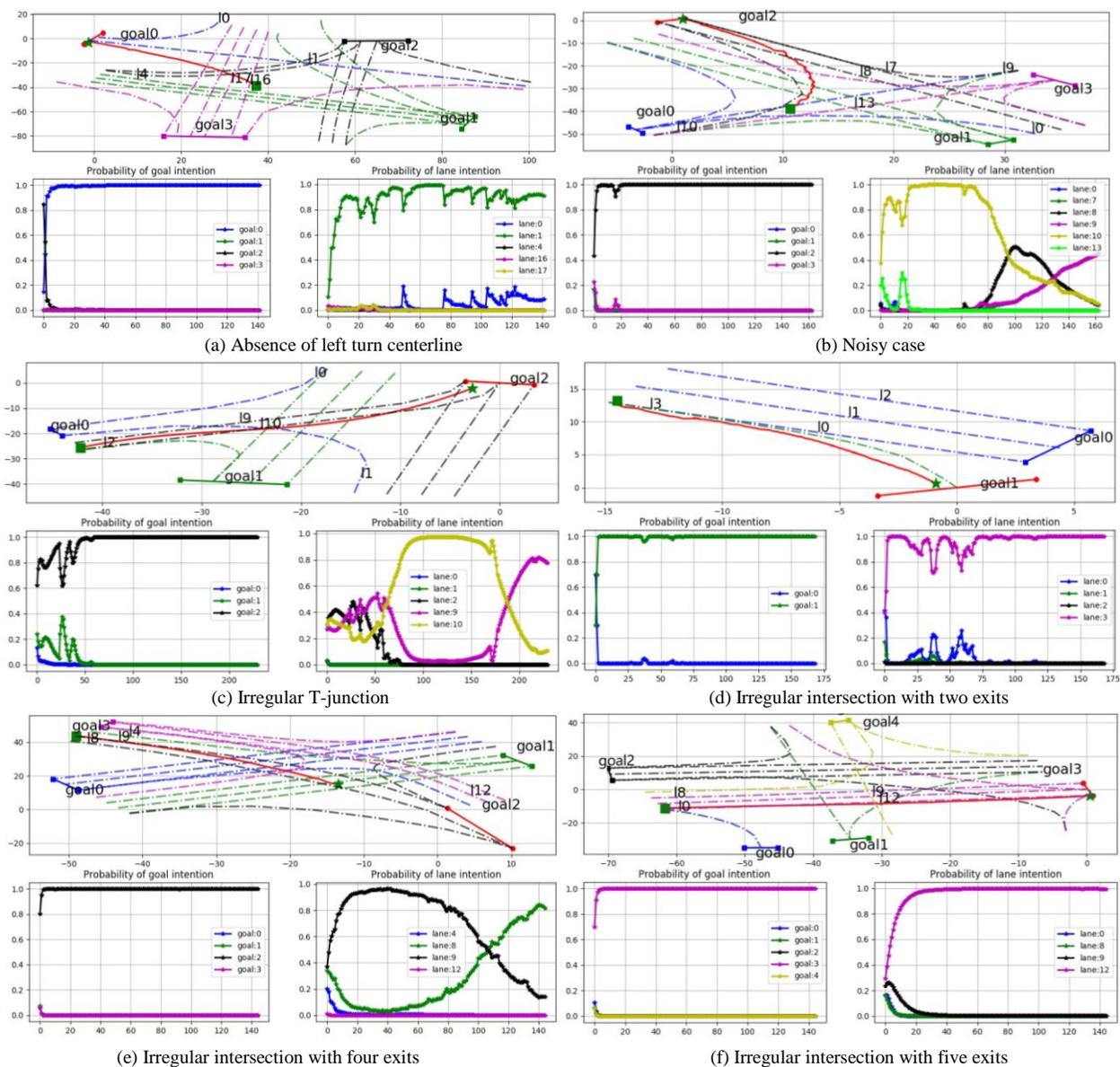

Figure 5. Qualitative results. Each dotted line represents centerline of each exit, strings at exit denote the index of goal intention, exit with red line is the actual goal intention, and strings at centerlines denote the index of virtual lane. Red line denotes target's trajectory, green squares and stars denote the start and end position. The last row are the probability distributions of goal intention and lane intention at each frame.

Linear Unit (ReLU) are used after each MLP, and the final layer has only one hidden unit. Both MLPs have two hidden layers, 128 units each. Lane interaction feature and goal interaction feature at current time stamp are the corresponding inputs of both MLPs. This model is trained in exactly the same way as the proposed model.

**Single Task Learning (STL):** This model is the same as the proposed model and outputs goal intention and lane intention, but it is trained for only goal intention prediction.

### C. Quantitative Results

The proposed model and baseline models are trained on the simulated training set, and evaluated on the simulated test set. The results are shown in Table II. When evaluating on the in-house dataset, we use the model pretrained on the simulated training set *without any fine tuning*. The predicted intention is always the one with the highest probability at each timestamp, and performance is evaluated by counting predictions of all frames. As we formulate open-set intention prediction as a matching problem, false positive is equal to false negative, so recall is equal to precision and only recall is reported for performance evaluation in this paper. Due to lack of lane intention label in in-house dataset, recall of lane intention prediction is only evaluated on the simulated test set. Recall of goal intention on in-house dataset is 84.6%, 93.7%, 94% and 95.9%. Considering the significant imbalance in the in-house dataset, recalls of goal intention in each case are shown in Table III. From results on the simulated test set and the in-house dataset, deep learning based models outperform KNN baseline. Compared with MLP baseline, MAAM improves the performance of intention prediction. Compared with STL baseline, the proposed model outperforms STL baseline in all cases. In-house dataset includes 264755 frames that are collected at unseen intersections, recalls of the proposed model at these unseen intersections are 0.952 and 0.947 for straight case and curved case respectively. High performance on the in-house dataset demonstrates that the proposed method can be generalized across different intersections, even at unseen intersections.

TABLE II. RECALL ON SIMULATED TEST SET

|  | KNN | MLP | STL | Ours |
|---|---|---|---|---|
| Goal intention | 0.869 | 0.935 | 0.966 | 0.988 |
| Lane intention | 0.721 | 0.883 | 0.913 | 0.965 |

TABLE III. RECALL OF GOAL INTENTION ON IN-HOUSE DATASET

|  | KNN | MLP | STL | Ours |
|---|---|---|---|---|
| Straight case | 0.875 | 0.961 | 0.953 | 0.971 |
| Curved case | 0.734 | 0.841 | 0.893 | 0.92 |

*D. Ablation Experiments*

TABLE IV. RESULTS OF ABLATION EXPERIMENTS ON THE IN-HOUSE DATASET

|  | Goal-only Model | Lane-only Model | L-GL Model | GL-L Model | Ours |
|---|---|---|---|---|---|
| Straight case | 0.92 | 0.951 | 0.97 | 0.972 | 0.971 |
| Curved case | 0.785 | 0.885 | 0.896 | 0.871 | 0.92 |

We conducted an ablation study to evaluate the importance of each module proposed in this paper, which is shown in Table IV. 1) Goal-only model: this model only predicts goal intention with goal interaction feature, no usage of any lane information; 2) Lane-only model: Lane attention and goal attention only use lane interaction feature; 3) L-GL model: L means that goal attention module only uses the representation $P_j^t$ of related virtual lanes, GL means that both goal and lane interaction features are used in lane attention module; 4) GL-L model: Both Goal and Lane interaction features are used in goal attention module, only Lane interaction feature is used in lane attention module. Without any lane level information, goal-only model perform well in straight case, while the performance is much worse in the curved case. This result shows that detailed context information plays a significant role in intention prediction. Compared with goal-only model, the performance of lane-only model is improved especially in the curved case by leveraging lane information. Adding auxiliary information either in goal attention module or lane attention module can improve the performance further. Full model with mutually auxiliary attention mechanism achieves the best performance. This confirms the effectiveness of mutually auxiliary attention mechanism.

*E. Qualitative Results*

Qualitative results on the in-house dataset regarding intersection layout variation are shown in Figure 5. In Figure 5 (a), the proposed model can predict the actual goal intention in the absence of left turn centerline. Furthermore, the proposed model can also predict accurately with noisy input as shown in Figure 5 (b). These two cases demonstrate the robustness to the noise in map information and target state. To show the generalization across different intersections, results at irregular T-junction, irregular intersections with two, four and five exits are shown in Figure 5 (c), (d), (e) and (f). When the target vehicle performs a lane change maneuver, the proposed model can not only predict the actual goal intention, but also the correct lane intention. In Figure 5 (c), (d) and (e), it is difficult to describe these behaviors with predefined closed-set intentions. These cases reveal that closed-set intention description of these behaviors are ambiguous, because closed-set intention is not strictly predefined in every possible scenario. In the intersection with 5 exits shown in Figure 5 (f), when a hypothetical target from left bottom corner is approaching goal0, it can be recognized as turning right under the closed-set intentions with going straight, turning left and turning right setup, but there are two exits corresponding to turning right intentions, this is a typical ambiguous case that closed-set intentions cannot handle. The proposed model can converge to the actual intention within only several frames and is very computationally efficient with only about 0.3ms running time for one target vehicle on machine with an Intel(R) Xeon(R) CPU E5 and 16G memory.

V. CONCLUSION

In this paper, we introduce open-set intentions to describe behaviors of human driver which can adapt to various intersections. Intention prediction is formulated as a matching problem between target vehicle and map elements. Under this formulation, the proposed model uses map-centric features for each open-set intention prediction instead of high dimensional rasterized images, experiments on the in-house dataset show that the proposed model can generalize across various intersection layout, even to unseen intersections, is computationally efficient with only 0.3ms running time for one target vehicle, and thus suitable to be deployed on real-world AD system. Furthermore, the proposed method can readily generalize to non-intersection area, with only one goal intention and several lane intentions to consider. Given goal intention and lane intention, learning based models also can be utilized to generate multi-modal future trajectories with respect to each possible intention, which will be left as future work.

REFERENCES

[1] Phillips, Derek J., Tim A. Wheeler, and Mykel J. Kochenderfer. "Generalizable intention prediction of human drivers at intersections." 2017 IEEE Intelligent Vehicles Symposium (IV). IEEE, 2017.
[2] Tang, Bo, Salman Khokhar, and Rakesh Gupta. "Turn prediction at generalized intersections." 2015 IEEE Intelligent Vehicles Symposium (IV). IEEE, 2015.
[3] Streubel, Thomas, and Karl Heinz Hoffmann. "Prediction of driver intended path at intersections." 2014 IEEE Intelligent Vehicles Symposium Proceedings. IEEE, 2014.
[4] Xu, Kecheng, et al. "Data driven prediction architecture for autonomous driving and its application on apollo platform." 2020 IEEE Intelligent Vehicles Symposium (IV). IEEE, 2020.

[5] Casas, Sergio, Wenjie Luo, and Raquel Urtasun. "Intentnet: Learning to predict intention from raw sensor data." Conference on Robot Learning. PMLR, 2018.

[6] Deo, Nachiket, and Mohan M. Trivedi. "Convolutional social pooling for vehicle trajectory prediction." Proceedings of the IEEE Conference on Computer Vision and Pattern Recognition Workshops. 2018.

[7] Chai, Yuning, et al. "Multipath: Multiple probabilistic anchor trajectory hypotheses for behavior prediction." arXiv preprint arXiv:1910.05449 (2019).

[8] Fang, Liangji, et al. "TPNet: Trajectory Proposal Network for Motion Prediction." Proceedings of the IEEE/CVF Conference on Computer Vision and Pattern Recognition. 2020.

[9] Veličković, Petar, et al. "Graph attention networks." arXiv preprint arXiv:1710.10903 (2017).

[10] Vaswani, Ashish, et al. "Attention is all you need." Advances in neural information processing systems. 2017.

[11] Sezer, Volkan, et al. "Towards autonomous navigation of unsignalized intersections under uncertainty of human driver intent." 2015 IEEE/RSJ International Conference on Intelligent Robots and Systems (IROS). IEEE, 2015.

[12] Kumar, Puneet, et al. "Learning-based approach for online lane change intention prediction." 2013 IEEE Intelligent Vehicles Symposium (IV). IEEE, 2013.

[13] Bahram, Mohammad, et al. "A combined model-and learning-based framework for interaction-aware maneuver prediction." IEEE Transactions on Intelligent Transportation Systems 17.6 (2016): 1538-1550.

[14] J. Colyar and J. Halkias, "Us highway 80 dataset," Federal Highway Administration (FHWA), Tech. Rep. FHWA-HRT-07-030, 2007.

[15] -----, "Us highway 101 dataset," Federal Highway Administration (FHWA), Tech. Rep. FHWA-HRT-07-030, 2007.

[16] Zyner, Alex, Stewart Worrall, and Eduardo Nebot. "Naturalistic driver intention and path prediction using recurrent neural networks." IEEE Transactions on Intelligent Transportation Systems 21.4 (2019): 1584-1594.

[17] Zyner, Alex, et al. "Long short term memory for driver intent prediction." 2017 IEEE Intelligent Vehicles Symposium (IV). IEEE, 2017.

[18] Zyner, Alex, Stewart Worrall, and Eduardo Nebot. "A recurrent neural network solution for predicting driver intention at unsignalized intersections." IEEE Robotics and Automation Letters 3.3 (2018): 1759-1764.